\pdfoutput=1
\documentclass[11pt]{article}

\usepackage[final]{acl}

\usepackage{times}
\usepackage{latexsym}

\usepackage[T1]{fontenc}
\usepackage[utf8]{inputenc}

\usepackage{microtype}

\usepackage{inconsolata}

\usepackage{graphicx}

\usepackage{mdframed}

\usepackage{multirow}

\title{Exploring the Potential of Large Language Models to Simulate Personality}

\author{
  Maria Molchanova\thanks{These authors contributed equally to this work.} \\
  MIPT, Russia \\
  \texttt{molchanova.ma.63@gmail.com}
  \And
  Anna Mikhailova\footnotemark[1] \\
  MIPT, Russia \\
  \texttt{anna.micj@gmail.com}
  \AND
  Anna Korzanova \\
  MIPT, Russia \\
  \texttt{korzanova.av@mipt.ru}
  \And
  Lidiia Ostyakova \\
  MIPT, Russia \\
  HSE University, Russia \\
  \texttt{ostyakova.ln@gmail.com}
  \And
  Alexandra Dolidze \\
  MIPT, Russia \\
  HSE University, Russia \\
  \texttt{dolidze.aa@mipt.ru}
}

\begin{document}
\maketitle

\begin{abstract}
% sentence like in intro = speaking like a human in terms of emotional reaction and behavior with the help of personality makes interaction more engaging
% what we really study = prompting LLM personality
% what we explored = models exhibit difference when answering the personality questionnaire and generating personality texts. High score of Neuroticism and low score of Agreeableness are hard to achieve. 
% what we provide = analytical framework

With the advancement of large language models (LLMs), the focus in Conversational AI has shifted from merely generating coherent and relevant responses to tackling more complex challenges, such as personalizing dialogue systems. In an effort to enhance user engagement, chatbots are often designed to mimic human behaviour, responding within a defined emotional spectrum and aligning to a set of values. In this paper, we aim to simulate personal traits according to the Big Five model with the use of LLMs. Our research showed that generating personality-related texts is still a challenging task for the models. As a result, we present a dataset of generated texts with the predefined Big Five characteristics and provide an analytical framework for testing LLMs on a simulation of personality skills. 

 % Despite the growing popularity of character-like AI assistants, response generation that accurately reflects specific personality features remains understudied.

\end{abstract}

\section{Introduction} %Аня М

% Текст пишу уставшая, поэтому надо проверить на ошибки, если будете заменять (Лида)
% Given the breakthrough of LLMs and their ability to generate coherent and contextually relevant responses, conversational agents are becoming increasingly capable of aligning with users' needs{ссылки}. In order to make human-machine interaction more engaging and intertaining, chatbots often impersonalize a specific character to foster an emotional bond between AI assistants and humans. Previous studies showed that imitating a personaly has a positive impact on user experience with dialogue systems by adding empathy in conversations{ссылка}, еще какие-то примеры. Personalization is widely used in different LLM-based applications --- > примеры про игроков and customer-oriented employees 
% Дальше нужно, по моему мнению, больше информации про сам промптинг. Какие есть способы задать промпты? Тоже нужно больше примеров и ссылок. Можно написать, что раньше в диалоговых системах использовали такие методы, чтобы задавать персону, а сейчас используют промпты. Написать, что тема совсем плохо изучена. Есть ли что-то, кроме BIG FIVE MODEL для моделирования персоны? Нет связи между введением и related work, из-за того что про промтинг почти нет информации. Может быть добавить ссылки на случаи, когда пытаются имитировать определенных персонажей или что-то подобное.  

After their breakthrough in NLP, large language models (LLMs) continue to improve, becoming integral to the modern digital world. Their ability to generate coherent and contextually relevant responses finds application in various fields, including dialogue systems~\cite{yi2024survey}.  

LLMs, with their ability to generate human-like text, open up new possibilities for conversational agents, making interactions with users more engaging. However, to boost this emotional connection between AI assistants and humans, dialogue systems need to be more personalized taking into account individual characteristics of  users. Prompting LLM-based multi-agent systems with personality can significantly enhance their capability to converse with empathy~\cite{rashkin2018towards}.

Imitating personality with LLMs may be a good way to develop usable AI assistants that perfectly align with users needs~\cite{tseng2024two}. Incorporating the desired character in assistants for each user can personalize user experience, prolonging dialogues and making them more lively. Thus, adjusting a bot personality to suit a user's emotional state can enhance user satisfaction with dialogue systems~\cite{prendinger2005using, polzin2000emotion}. In LLM-based multi-agent systems, agents can take on different roles with description of their personality. For example, such agents can be applied to the gaming industry, portraying players with different characteristics, or in business applications, where agents can take on the roles of customer-oriented employees to provide guidance in certain fields of expertise~\cite{guo2024large}.

The current study aims at providing evidence on how different LLMs may be prompted with Big Five personality model, and how consistent they are in the imitated personalities. We explore not only how well LLMs associate personality traits with corresponding behavior, but also their knowledge about language patterns and life goals related to the prompted personality. As a result of our study we develop an open-source analytical framework to encourage further research in personality simulation by LLMs.

% Чтобы расширить introduction, можно прописать списком contributions. Мы сделали это, то и третье. 

\section{Related Work}
\paragraph{Big Five Personality Model} 

The Big Five personality model is considered to be a valid approach to evaluate one's personality traits due to its reliability and practicality~\cite{john2008paradigm}. Compared to MBTI personality measurement, the Big Five model has proven its validity across different populations~\cite{schweiger1985measuring}. According to the Big Five model, an individual's personality can be represented along five dimensions, namely Extraversion, Neuroticism, Agreeableness, Conscientiousness, and Openness to Experience. Each dimension is a dichotomy with low and high trait manifestations. Recent research adds a new dimension, Honesty-Humility, to the original model~\cite{costa2008revised}.

Initially, the Big Five model is based on the linguistic hypothesis, claiming that the natural language descriptors of person's traits, moods, and characteristics can be generalized and represented as a personality model. Subsequently, there appear also psychological questionnaires, namely those as TDA \cite{goldberg1992development}, NEO \cite{costa1978objective}, and BFI \cite{john1991big}, that help to determine personality traits by answering a predefined set of questions. All the questionnaires show respectable validity and reliability, but BFI questionnaire stands out for question formulation~\cite{john2008paradigm}. The Big Five model is critiqued for lacking detailed personality classification, as all traits seem to be abstract~\cite{john2008paradigm}; moreover, the model only includes traits linked to natural language analysis, neglecting non-social personal traits~\cite{trofimova2014observer}. Nevertheless, these characteristics can be advantageous in the context of LLMs, as they can be utilized to generate a personality in natural language modeling using the Big Five model.

% Это основной для понимания статьи абзац, но из него вообще ничего не ясно. Нужно больше подробностей. Если успеется, то нарисовать бы схему. (Лида)

\paragraph{Personality of Large Language Models}
%вводная часть про LLM (следуют инструкциям, могут принимать роли и т.д.)
%LLM + генерация персональности (это обзор статей)

The research devoted to personality in LLMs focuses on two main aspects, including internal LLM personality and LLM-prompted personality. LLM personality is usually analyzed by leveraging the Big Five questionnaire results, LLM-generated text analysis (both linguistic and psychological), or automated text classification for the Big Five traits.

Studies on LLM personalities, e.g., \cite{sorokovikova2024llms}, provide statistics on how LLMs answer the different the Big Five questionnaires (IPIP-NEO, BFI-44). This should reveal the LLM hidden personality. The findings seem to be quiet impressive, as they portray GPT-4 and Mixtral as open, agreeable, conscientious, extroverted and emotionally stable agents capable of engaging conversations. However, other studies disagree with this conclusion, as the questionnaire results do not seem to reflect real model personality and do not help to predict model actions~\cite{ai2024cognition}.

Studies on how we can induce personality in LLMs demonstrate different ways a personality can be prompted to an LLM. Some approaches suggest describing personality based on the personal facts (e.g., providing the LLM with a personality description from PersonaChat dataset) or based on descriptions of each the Big Five characteristic~\cite{serapiogarcía2023personality, liu2024dynamic}. The other research also tests personality simulation through text generation~\cite{jiang2024evaluating} or sentence completion~\cite{hilliard2024eliciting}. Personality modeling is usually affected by the model size, as it is stated in all above-described studies. 
% 
% Этот параграф лучше, но формулировки предложений очень сложные (много доп конструкций внутри предложений). Я бы посоветовала упростить текст, чтобы его было проще читать, и немного разнообразить предложения (Лида)
\section{Methods} 

Based on previous research, LLM personality prompting should be approached from two perspectives, including personality questionnaire answering and personal text generation. Therefore, the executed LLM personality analysis consists of two stages. First, the personality questionnaire is used to evaluate the LLM's understanding of the links between personality traits and the associated behaviors. Second, LLMs are prompted to generate texts regarding their induced personality. Compared to the first analysis stage, personality-related text generation requires from an LLM knowledge not only about personality behavior but also about personal language habits, life goals, and inclinations. LLM-generated texts are analyzed in different ways to build a complete picture of the LLM's personality simulation abilities that involves human evaluation of the generated texts, automatic evaluation, and linguistic feature analysis. The analysis is conducted among the proprietary LLMs, including GPT-3.5 Turbo and GPT-4 Omni~\footnote{\url{https://platform.openai.com/docs/models}}, Mixtral 8x22B~\footnote{\url{https://mistral.ai/news/mixtral-8x22b/}}, Claude 3 Haiku~\footnote{\url{https://www.anthropic.com/news/claude-3-haiku}}. These models were chosen because of their speed, affordability, comparative input context length, and performance.

\section{LLMs Answering the Big Five Questionnaire} 

Testing prompted LLM personality through questionnaire was based on the BFI-44 questionnaire due to its size and reliability. Each item in the questionnaire corresponds to one of the Big Five traits. So, the model was given instructions to simulate high or low trait score and answer the questionnaire part related to this trait (see in Appendix\ref{app:prompt_example_1}). Then the results were aggregated and the final trait score was revealed.

\begin{figure*}[ht!]
    \centering
    \includegraphics[width=1.8\columnwidth]{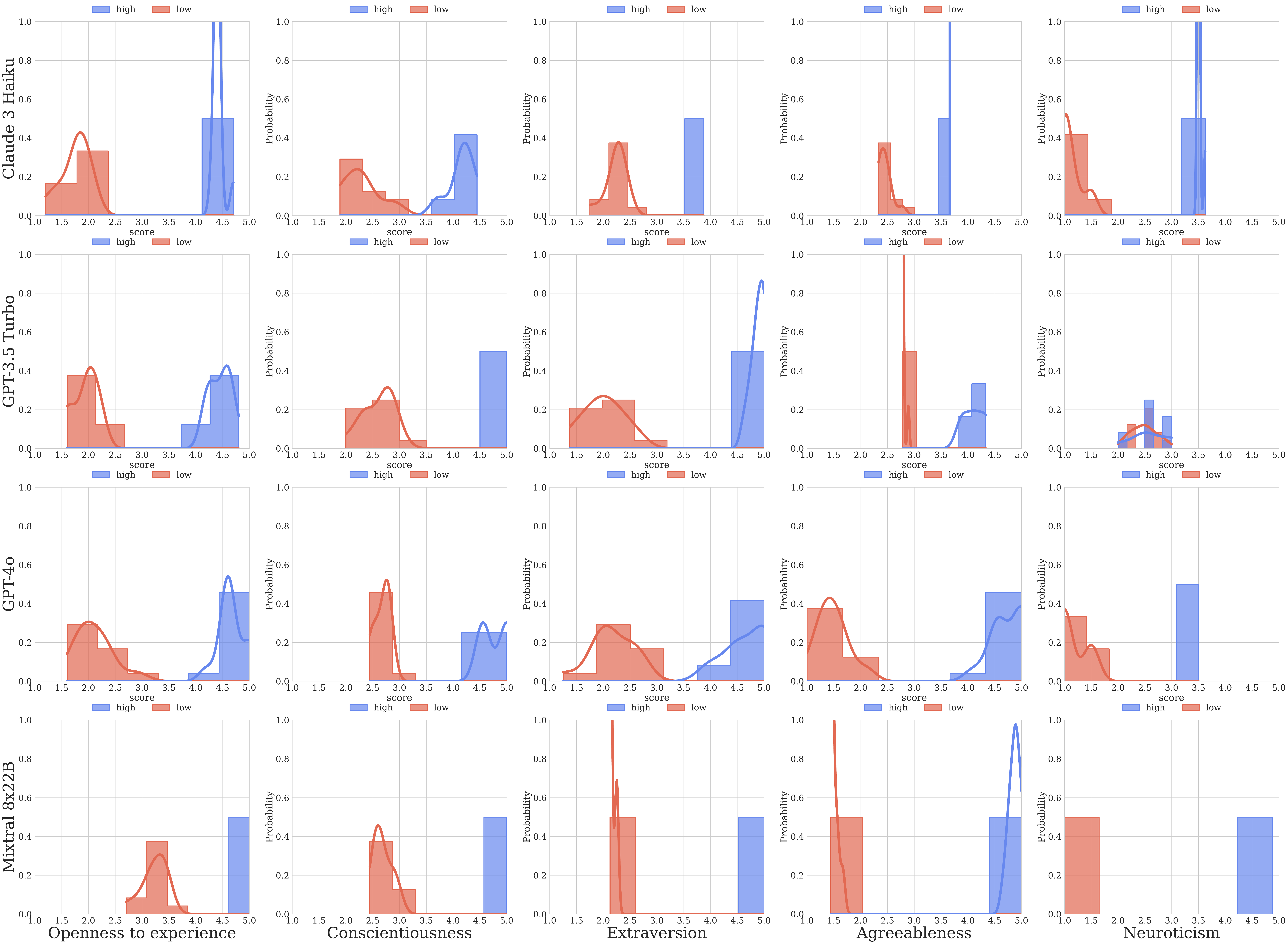}    \caption{Distribution of BFI-44 trait scores for different LLMs prompted to simulate high or low Big Five traits.} % Подпись к изображению
    \label{fig:bfi-results} % Метка для ссылки на изображение в тексте
\end{figure*}

Following ~\cite{serapiogarcía2023personality}'s approach to questionnaire answering, we also measured the reliability and validity of the LLM's questionnaire answers. Reliability tests included measuring Cronbach's alpha and Guttman's lambda (see Appendix \ref{sec:appendix-reliability-test}). As the reliability metrics exceed 0.85, the questionnaire answers of LLMs may be considered real and plausible. 

To observe more precisely how the model's answers differ regarding high or low trait prompting, a distribution map was built (see Figure~\ref{fig:bfi-results}). The map reveals that none of the observed models demonstrates consistent personality generation when assessing the personality-related statements. However, some models show accurate performance considering particular traits. So, the Claude model seems to have inner knowledge about such traits as Openness to experience and Conscientiousness, whereas the Mixtral model about Agreeableness and Neuroticism. The GPT models are able to differentiate between high and low levels of Extraversion; additionally, the GPT-4 Omni model excels at distinguishing high and low levels of Agreeableness.

\section{LLMs Generating Big Five Personality Texts} 

\subsection{Experiment Settings}

In order to evaluate LLM simulation of a prompted personality in generated texts, a set of questions was prepared. The questions concerned the LLM preferences, perspectives, and related matters. The prompt for generation consisted of specified trait scores, a description of each trait, and the chosen question (see Appendix \ref{app:prompt_example_2}). Two categories of data were generated:
\begin{itemize}
    \item Texts generated using only one trait. The purpose of this task is to evaluate the model's ability to imitate a specific characteristic.
    \item Texts generated using personality settings that included every factor of the Big Five model, in order to examine the overall simulation of traits. The analysis of this second type will be addressed in future research.
\end{itemize}

To generate responses reflective of individual traits, we tested all possible discrete score variations from 1 to 5 to ensure the diversity of generated texts and adjusted the temperature settings to 0, 0.5, 0.7, and 0.9. To analyze the LLM responses to multiple personality traits, we used data points formed by a normal distribution. The mean and variance for the distribution of each trait were calculated by analyzing a dataset of 1,015,342 questionnaire answers~\cite{oceandataset} that contained the personality traits information of actual users. These data were also obtained using different temperature settings.

The Claude model exhibited p performance in deciding whether to deliver a direct answer to a question or to provide a recommendation consistent with the designated personality profile. Additionally, the model sometimes explicitly disclosed its own personality traits despite instructions in the prompt to avoid doing so. In such instances, responses were edited by masking direct references to personality characteristics or by removing content that did not align with the task objectives.

\subsection{Human Evaluation}

% Total amount of about 288 generated text samples is split into batches of 20 texts per batch for human evaluation.

From the LLM-generated data, there were selected about 288 text samples for human annotation in total. The amount of texts is equally distributed across all the models under analysis and balanced by model temperature. The data is equally split into batches with 20 texts per batch for an annotator.

For human evaluation, we recruited 8 annotators to infer the personality traits in the texts generated by the LLMs. They were provided with the trait descriptions and a detailed instruction, explaining required steps to complete the annotation. Annotators scored each text on a scale from -2 to +2, depending on the intensity each personality trait is expressed, where -2 is low score, 0 is neutral, and +2 is high score. From a given list of reasons, raters also chose why they thought the trait should be scored that low or high. To extract linguistic patterns from the texts, annotators were asked to highlight words and word phrases that made them think the person had a particular personality trait.

As perceptions of psychological aspects of  personality can vary among individuals, each models' answer was assigned to 3 different annotators, and the agreement between them was evaluated using Fleiss' kappa. The final assessments were acquired via voting. Inner-annotation agreement metrics are displayed in Table~\ref{tab:IIA}. We conducted metric calculations at multiple levels, including identifying the existence or absence of a trait in the text, and calculating the score group (non-distinguishable, low, middle, high).

\begin{table}[ht!]
    \centering
    \begin{tabular}{l|c|c}
\hline
\textbf{Trait} & \textbf{Level-1} & \textbf{Level-2}\\
\hline
         Openness to experience&  0.67 & 0.69\\
         Conscientiousness&  0.66&0.66\\
         Extroversion&  0.62&0.58\\
         Agreeableness& 0.71&0.70\\
         Neuroticism&  0.59&0.57\\
         \hline
    \end{tabular}
    \caption{Fleiss' kappa for human annotation results}
    \label{tab:IIA}
\end{table}

\subsection{Automatic Evaluation}

\begin{table*}[ht!]
    \centering
    \begin{tabular}{l|c|c|c}
    \hline
    \multicolumn{4}{c}{\textbf{Level 1}} \\
    \hline
    \textbf{Trait} & \textbf{Weighted Precision} & \textbf{Weighted Recall} & \textbf{Weighted F1 score} \\
    \hline
Openness to experience & 0.88 & 0.88 & 0.87 \\
Conscientiousness & 0.80 & 0.80 & 0.80 \\
Extraversion & 0.81 & 0.76 & 0.76 \\
Agreeableness & 0.75 & 0.74 & 0.71 \\
Neuroticism & 0.64 & 0.63 & 0.63 \\
    \hline
    \end{tabular}
    \vspace{0.5cm} % Add vertical space between tables
    \begin{tabular}{l|c|c|c}
    \hline
    \multicolumn{4}{c}{\textbf{Level 2}} \\
    \hline
    \textbf{Trait} & \textbf{Weighted Precision} & \textbf{Weighted Recall} & \textbf{Weighted F1 score} \\
    \hline
Openness to experience & 0.79 & 0.80 & 0.78 \\
Conscientiousness & 0.75 & 0.75 & 0.75 \\
Extraversion & 0.71 & 0.64 & 0.65 \\
Agreeableness & 0.70 & 0.71 & 0.67 \\
Neuroticism & 0.52 & 0.53 & 0.50 \\
    \hline
    \end{tabular}
    \caption{LLM-based classifier performance at different levels.}
    \label{tab:classifier_similarity_level_1_2}
\end{table*}

To extrapolate the human evaluation of LLM-simulated personality and to enable future automated evaluation for other models and prompts, we created an LLM-based personality classifier. This classifier is based on the CARP method~\cite{sun2023text}, which initially prompts the LLM to provide a list of clues (i.e., keywords, phrases, contextual information, semantic relations, semantic meanings, tones, references) that indicate a personality trait. The model subsequently analyzes the extracted data and, based on this analysis, detects the personality trait, providing an explanation for its response. The instructions for personality detection given to the model were exactly the same as those provided to the human annotator (see Appendix \ref{app:prompt_example3}). The GPT-4 Omni model was chosen as the base classifier due to its superior performance compared to other LLMs.

The analysis of the correspondence between classifier evaluations and those of human annotators includes several levels:
\begin{itemize}
    \item \textbf{Level 1}: Evaluating the presence or absence of trait markers in the text. The classifier's and annotators' responses are considered consistent if both determine that the markers are either present or absent.
    \item \textbf{Level 2}: Categorizing the scores into four groups. The Low group includes scores of -2 and -1. A score of 0 represents a balanced demonstration of the trait and is classified into a distinct category. The High group contains scores 1 and 2. The Non-distinguishable category is for cases where the trait cannot be identified. The classifier's results and the human annotators' results are compared among these groups.
    \item \textbf{Level 3}: Analyzing the differences between the classifier's evaluations and the average scores provided by the annotators.
\end{itemize}

Table~\ref{tab:classifier_similarity_level_1_2} presents the examination of the similarity between the LLM-based classifier evaluations and the human annotators' responses at Levels 1 and 2. The Level 3 metrics are displayed in Table~\ref{tab:classifier_similarity_level_3}.

\begin{table}[ht!]
    \centering
    \begin{tabular}{l|c}
\hline
\textbf{Trait} & \textbf{MAE}\\
\hline
         Openness to experience&  0.44\\
         Conscientiousness&  1.05\\
         Extroversion&  0.88\\
         Agreeableness& 0.54\\
         Neuroticism&  1.77\\
         \hline
    \end{tabular}
    \caption{LLM-based classifier, level 3.}
    \label{tab:classifier_similarity_level_3}
\end{table}

The most problematic trait for the classifier is Neuroticism, which also has a low inner-annotation agreement metric within the group.

\begin{figure*}[ht!]
    \centering
    \includegraphics[width=1.8\columnwidth]{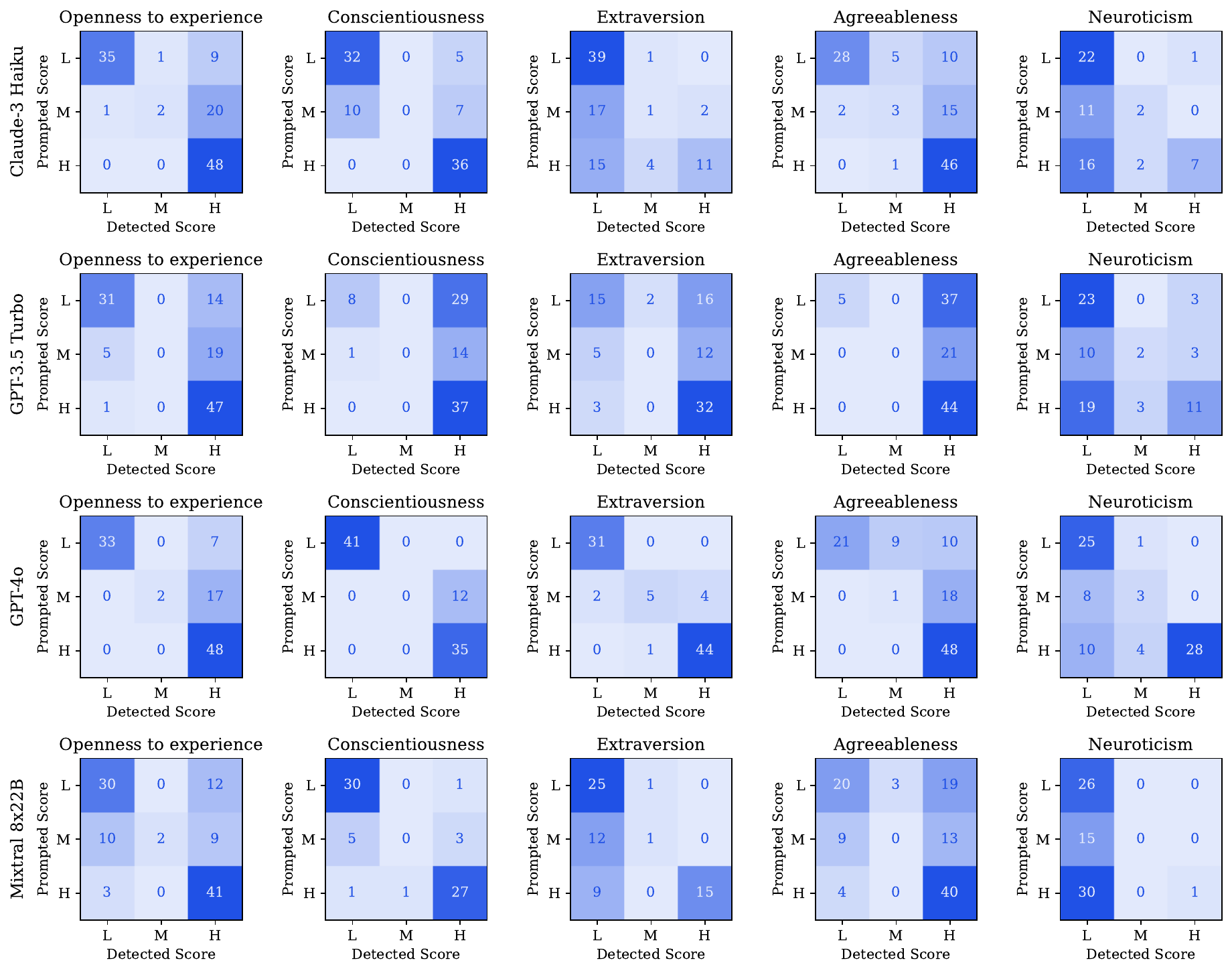}
    \caption{Confusion matrix for prompted and detected trait level. L stands for low trait level, M for middle level and H for high level.} % Подпись к изображению
    \label{fig:cm_combiled_anole} % Метка для ссылки на изображение в тексте
\end{figure*}

Figure~\ref{fig:distribution_trait_detection} presents the results of the classifier's detection of the trait included in the prompt and shows how identifiable the presence of the prompted trait is in the generated text. Figure~\ref{fig:cm_combiled_anole} demonstrates the distribution of trait scores in generated texts when the model failed to recognize the personality. The confusion matrices in Figure~\ref{fig:cm_combiled_anole} illustrate how accurately the prompted score groups for the traits were identified by the classifier and what types of errors were made.

The results of the Big Five classifier on generated text data show that
among all the models analyzed, the trait of Openness to Experience was consistently simulated throughout the text generation process. The models demonstrated a robust understanding of the prompts, effectively capturing and conveying both low and high scores in the generated texts. Although the presence of Agreeableness trait was often detected in the generated texts, several models mistakenly represented high score of the trait personality, even if the prompt required low trait score.

The models had significant difficulty simulating Neuroticism. Nearly half of the generated texts did not consist of any detectable trait markers. The majority of models consistently presented personality with low scores of Neuroticism, even when the high scores were prompted. The results for Extraversion and Agreeableness exhibit notable variations among different models. In almost all cases, when a middle score for a trait was not correctly represented, the prompted personality was frequently identified as either a low or high score of the trait. 

Additionally, we found that when the model exhibits an identifiable bias towards either high or low scores, the generated texts with a middle score of a trait would frequently follow this bias. As a result, using the base method of prompting a middle score of a trait does not produce the desired outcome. 

The comparison between models reveals that:

\begin{itemize}
    \item \textbf{Claude-3 Haiku} model demonstrated a robust ability for imitating Openness to Experience: texts generated with this trait in the prompt can be easily identified for its existence, and the model successfully simulates both low and high scores. The presence of the Conscientiousness trait was detected in 75\% of cases, and the scores were reproduced with high precision in the texts. The model is capable of simulating different scores of Agreeableness; however, with slightly less precision. The model demonstrated strong bias toward low scores for Extraversion and Neuroticism, regardless of the prompt settings.
    \item \textbf{GPT-3.5 Turbo} model was capable of imitating Openness to Experience in generated texts, despite some flaws. In addition, the model demonstrated significant bias towards Conscientiousness (high score), Agreeableness (high score), and middle bias towards Neuroticism (low score). In almost half of the cases, it demonstrates a high score for Extraversion, even when low scores are requested in the prompt.
    \item \textbf{GPT-4 Omni} model demonstrated a strong ability to imitate the Big Five traits such as Openness to Experience, Conscientiousness, and Extraversion. Regarding the trait Agreeableness, the model had a moderate bias towards high scores. The model successfully simulated various scores of Neuroticism in the texts; however, it had a slight bias towards expressing low scores and a tendency to under-demonstrate the trait in texts where a low score was specified in the prompt.
    \item \textbf{Mixtral 8x22B} model, along with other models, is capable of imitating the trait Openness to Experience, however with a few mistakes. Although the model could not demonstrate the Conscientiousness trait in a significant number of cases (43\%), successfully simulated both low and high personality scores in detected cases. The model also demonstrates a minor bias when generating low scores of Extraversion and high scores of Agreeableness, together with a strong bias towards low scores of Neuroticism.
\end{itemize}

\subsection{Linguistic Features Evaluation}

\begin{figure*}[ht!]
    \centering
    \includegraphics[width=1.6\columnwidth]{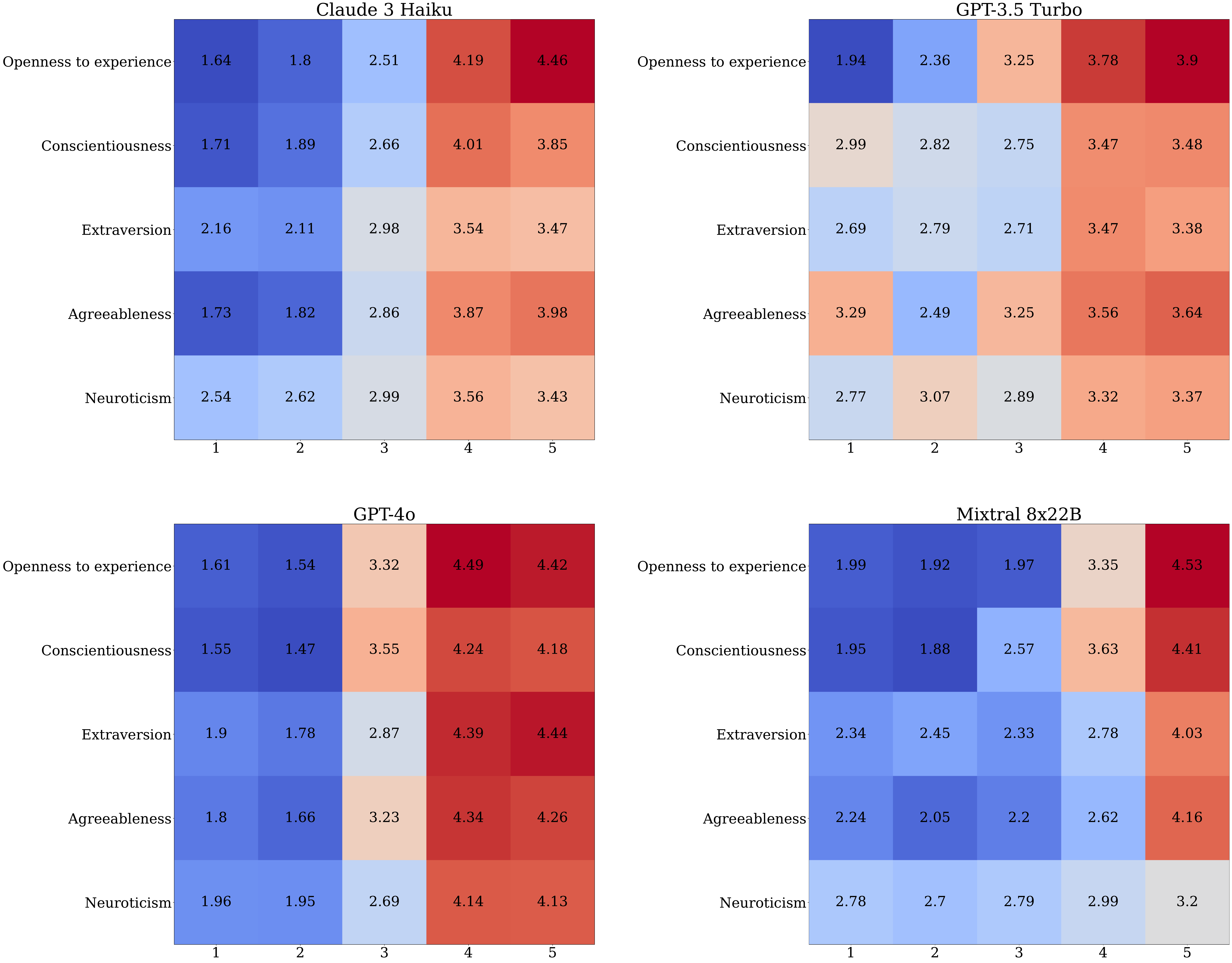}
    \caption{Averaged trait-level scores of the 5 most similar LLM-generated texts, grouped by the Big Five traits.} % Подпись к изображению
    \label{fig:cosine-similarity-results} % Метка для ссылки на изображение в тексте
\end{figure*}

At last the generated texts were analyzed linguistically. First, the texts prompted with different trait scores were compared to each other based on vocabulary used. This required to measure similarity between texts with different trait scores. The simplest way to assess text similarity is the use of TF-IDF vectorization\footnote{\url{https://scikit-learn.org/stable/modules/generated/sklearn.feature_extraction.text.TfidfVectorizer.html}} for comparing generated texts. The heatmap shown in Figure \ref{fig:cosine-similarity-results} displays the mean scores of the chosen Big Five traits for the five most similar texts to a given text (the scores are averaged among texts with the same trait; similar texts with other traits are not considered). The results show that Claude Haiku and GPT-4 Omni generate the most lexically consistent texts, which at the same time differ according to the level of trait manifestation.

In addition to that, we analyzed the extracted text samples assigned to the traits by annotators. Personality types are usually indicated by the usage of certain lexicons in the texts~\cite{ashton2004hierarchical}. As LLMs are able to align with human linguistic patterns simulating personality types~\cite{jiang2023personallm}, we examined lexical choices in the generated texts across different trait scores. 

Text samples were preprocessed and then lemmatized for further analysis with the use of the spaCy library~\footnote{\url{https://spacy.io/}}. The extracted lemmas were categorized by a part of speech and the trait scores specified in prompts. The most common verbs, nouns, and adjectives depicting personality features, according to annotators' opinions, are presented in Appendix~\ref{lexicon}. 

Nouns and adjectives more effectively represent the spectrum of the five personality traits compared to verbs. Analysis revealed that 16\% of the linguistic patterns in generated texts were derived from prompts.
Texts with high or low scores exhibit more noticeable variations in lexicons, whereas texts with a neutral trait score (equal to 0) tend to display linguistic patterns that are common to both high and low-scored texts. With the exception of Neuroticism, the lexicons of zero-scored samples are more similar to those that have low scores, as they share up to 34\% of patterns. The distribution of lemmas across trait scores demonstrates that employing binary scores for persona definition in LLM prompts is more reasonable, as most models struggle to capture subtle nuances of personality features.

% neutral analysis
% NEUROTICISM
% INTERSECTION WITH LOW SCORES: 0.11
% INTESECTION WITH HIGH SCORES: 0.13
% EXTRAVERSION
% INTERSECTION WITH LOW SCORES: 0.30
% INTESECTION WITH HIGH SCORES: 0.21
% OPENNESS
% INTERSECTION WITH LOW SCORES: 0.34
% INTESECTION WITH HIGH SCORES: 0.23
% CONSCIENTIOUSNESS
% INTERSECTION WITH LOW SCORES: 0.28
% INTESECTION WITH HIGH SCORES: 0.16
% AGREEABLENESS
% INTERSECTION WITH LOW SCORES: 0.26
% INTESECTION WITH HIGH SCORES: 0.17

\section{Analytical Framework}

In our exploration of LLM capacity to generate personality-consistent content, we developed and released an analytical framework that empowers researchers to replicate research with custom parameters. Our customizable framework allows users to integrate their own models and configurations, and to modify prompts for personality simulation. It supports the incorporation of different personality questionnaires beyond the BFI-44 used in our study. Additionally, users can add custom questions for the models to generate responses to.

Within this framework, the evaluated models respond to items from the chosen questionnaires, and their answers are used for graphical analysis The models also generate texts based on specified prompts, which are then automatically analyzed using the LLM-based classifier. This process assesses the accuracy and consistency with which the models adhere to the assigned personality traits during text generation. By making this framework publicly available on GitHub~\footnote{\url{https://github.com/mary-silence/simulating\_personality}}, we aim to contribute a valuable tool for advancing research in personality simulation by LLMs. The dataset of generated texts used in this research is also available at the same link.

\section{Discussion}

% Neuroticism самая сложная для симуляции черта, возможно ее вообще не стоит использовать
% Текстовый анализ дает лучшее представление о симуляции персональности, чем опросники, хоть и сложнее в исполнении

% 1. What has shown its efficiency? Questionnaires or text generation?
% 2. What we can learn from automatic LLM evaluation?
% 3. Does it reflect real data quality? What linguistic analysis shows?
Leveraging both questionnaire answering and text generation for testing LLM personality simulation skills gives a complex picture how LLM can answer certain personality-related statements and formulate statements about itself. Seemingly, the text generation is more informative regarding the Big Five personality simulation, as it is originally based on the linguistic hypothesis and is associated with the use of specific vocabulary. The complexity of text analysis process may be overcome when using modern NLP tools and LLM power. 

Concerning modern LLM performance on personality simulation, Neuroticism, among all the Big Five traits, is the most challenging trait to simulate by LLMs, as it was not detected in the questionnaire answers or by human analysis. The explanation might be that high score of Neuroticism is mostly connected to direct actions and ability to demonstrate emotional response, and LLMs cannot be proactive or show their emotions. Therefore, when prompting personality to an LLM, Neuroticism is better to set neutral or to avoid mentioning. However, overall LLM performance may be estimated as good, because LLMs demonstrate knowing of Big Five trait characteristics and seem to differentiate between various levels of trait manifestation.

\section{Conclusion and Future Work}

In this paper, we explored the potential of LLMs to simulate personality traits across different types of tasks. The results of personality simulation varied significantly depending on the task type. For example, when an LLM was asked to response questionnaire aligned with its prompted personality, the model generally performed well, though occasionally exhibited minor bias. However, when generating texts, LLM behavior differed significantly. In some cases, the models were unsuccessful in effectively conveying the presence of personality traits with recognizable scores in the generated text. Particularly, for the Big Five traits such as Agreeableness and Neuroticism, some models demonstrated strong bias towards certain score groups, even when explicitly prompted to simulate opposing trait scores. Thus, the model's behavior in responding to questionnaires and in generating texts may have been inconsistent. This difference may be related to the models' inherent default role, where high Agreeableness and low Neuroticism scores are considered essential for being an effective AI assistant. 

Therefore, when generating text that encompasses the full spectrum of personality trait scores according to the Big Five model, it is crucial to assess whether the model exhibits any bias towards specific trait scores. To facilitate further research on the capabilities of LLMs in simulating personality and conducting prompt engineering experiments for personality assignment, we have released a repository containing the analytical framework developed for this study.

Future research should focus on detailed multiple-trait personality generation, as AI models might be capable of incorporating all five traits of the Big Five model into the agent's personality. It is appealing due to the perspective of creating complex agent characters or game characters. 

\section*{Limitations}

% Theoretical limitation is use of Big Five model despite 4 premises of it.
% Analytical limitation includes not dealing with different prompt types. 

Limitations of the current study include using the Big Five personality model despite its assumptions~\cite{costa1999five} about proactivity and variability of a research subject. Besides, the effect of LLM size and training data was not taken into account as it may affect how models are capable of answer the questionnaire and generate texts. The study is also limited by usage of only partial human dataset annotation.

\section*{Acknowledgements}
This work was supported by a grant for research centers in the field of artificial intelligence, provided by the Analytical Center for the Government of the Russian Federation in accordance with the subsidy agreement (agreement identifier 000000D730321P5Q0002) and the agreement with the Moscow Institute of Physics and Technology dated November 1, 2021 No. 70-2021-00138.

%\bibliographystyle{acl_natbib}

\newpage
\appendix

\section{Appendix}
\label{sec:appendix-reliability-test}

\begin{table}[ht!]
\begin{center}
\begin{tabular}{l|c|c}
\hline
\textbf{OCEAN trait} & \begin{tabular}[c]{@{}c@{}}\textbf{Cronbach's} \\ \textbf{alpha} \end{tabular} & \begin{tabular}[c]{@{}c@{}}\textbf{Guttman's} \\ \textbf{lambda} \end{tabular}\\
\hline
Neuroticism & 0.87 & 0.96\\
Agreeableness & 0.87 & 0.97\\
Extroversion & 0.94 & 0.96\\
Conscientiousness & 0.88 & 0.95\\
Openness & 0.94 & 0.96\\
to experience & & \\
\hline
\end{tabular}
\caption{Reliability measurements among Big Five traits.}\label{table:reliability-scores}
\end{center}
\end{table}

\section{Appendix}
\begin{figure}[hp!]
    \centering
    \includegraphics[width=1\columnwidth]{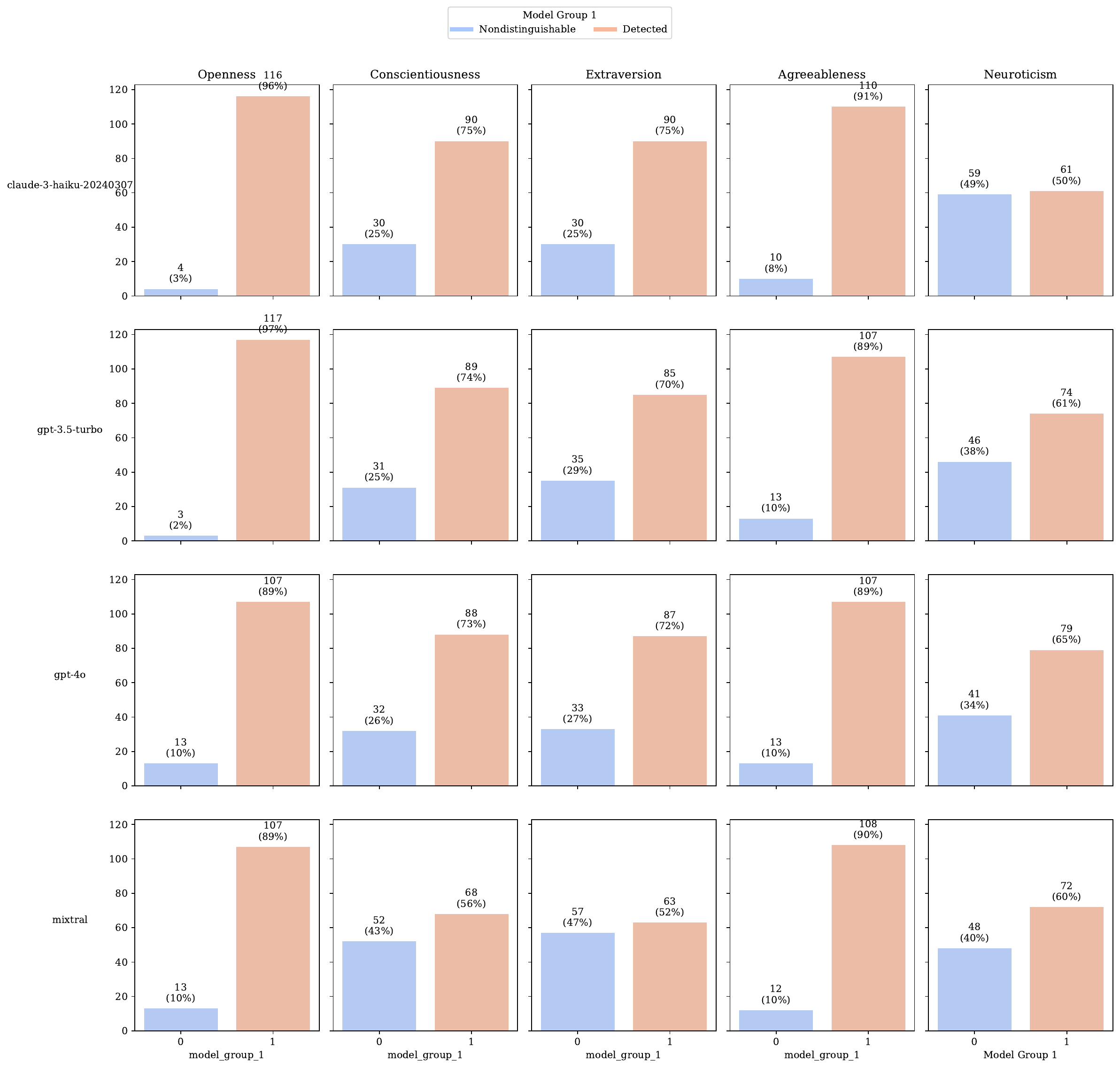}
    \caption{Distribution of texts where trait were non-distinguishable and clear to detect.} % Подпись к изображению
    \label{fig:distribution_trait_detection} % Метка для ссылки на изображение в тексте
\end{figure}

\begin{figure}[ht!]
    \centering
    \includegraphics[width=1\columnwidth]{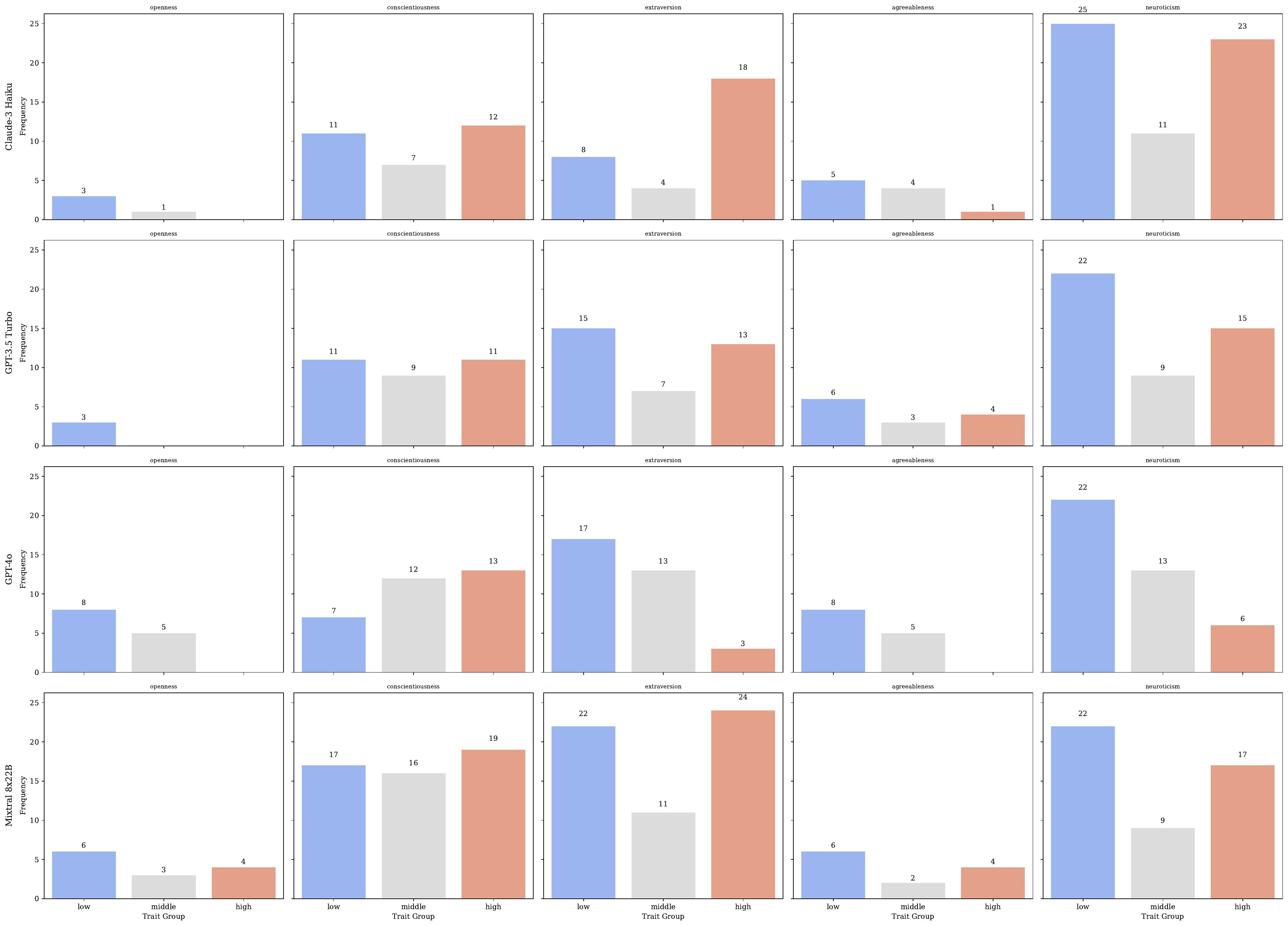}
    \caption{Distribution of the prompted personality level in texts claimed non-distinguishable for trait detection.} % Подпись к изображению
    \label{fig:nd-distribution} % Метка для ссылки на изображение в тексте
\end{figure}

% This is a section in the appendix.

\section{Appendix}
\label{sec:prompts}

\subsection{Prompt}

\label{app:prompt_example_1}
\mdfsetup{%
   middlelinecolor=black,
   middlelinewidth=1pt,
   backgroundcolor=gray!10,
   roundcorner=5pt,
   frametitle={Prompt for Questionnaire},
   frametitlerule=true
}
   
\begin{mdframed}
% [linecolor=gray, backgroundcolor=white, hidealllines, width=0.5]
% \begin{verbatim}
System prompt:\\
TASK:\\
Indicate your level of agreement or disagreement with the statement in the CHARACTERISTICS according to your PERSONALITY. Use only the PROVIDED OPTIONS.\\
\\
PERSONALITY:\\
```\\
$\bigr[$ TRAIT PROMPT $\bigr]$\\
```\\
\\
PROVIDED OPTIONS:\\
- disagree strongly with the statement\\
- disagree a little with the statement\\
- agree nor disagree with the statement\\
- agree a little with the statement\\
- agree strongly with the statement\\
\\
Provide your output only from the constant list ['disagree strongly with the statement', 'disagree a little with the statement', 'agree nor disagree with the statement', 'agree a little with the statement', 'agree strongly with the statement'] without explanation.\\
\\
User prompt:\\
CHARACTERISTICS:\\
```\\
$\bigr[$ STATEMENT $\bigr]$\\
```
\end{mdframed}

% Переход на новую страницу
\subsection{Prompt}

\label{app:prompt_example_2}
\mdfsetup{%
   middlelinecolor=black,
   middlelinewidth=1pt,
   backgroundcolor=gray!10,
   roundcorner=5pt,
   frametitle={Prompt for text generation},
   frametitlerule=true
}
   
\begin{mdframed}
% [linecolor=gray, backgroundcolor=white, hidealllines, width=0.5]
% \begin{verbatim}
System prompt:\\
TASK:\\
Answer the QUESTION according to your PERSONALITY. Use INSTRUCTION. Use at most 5 sentences. Do not mention your personality traits in the text. Type only the answer, without the information about your personality score.\\
\\
PERSONALITY:\\
- Your personality trait $\bigr[$ TRAIT $\bigr]$ is rated as \\$\bigr[$ SCORE $\bigr]$.\\
\\
INSTRUCTION:\\
- The personality trait is rated from 1 to 5. 1 is the lowest score and 5 is the highest score.\\
- $\bigr[$ DEFINITION OF LOW SCORE $\bigr]$\\
- $\bigr[$ DEFINITION OF HIGH SCORE $\bigr]$\\
\\
User prompt:\\
QUESTION:\\
```\\
$\bigr[$ QUESTION $\bigr]$
```
\end{mdframed}

\subsection{Prompt}

\label{app:prompt_example3}
\mdfsetup{%
   middlelinecolor=black,
   middlelinewidth=1pt,
   backgroundcolor=gray!10,
   roundcorner=5pt,
   frametitle={Prompt for LLM-based classifier},
   frametitlerule=true
}
   
\begin{mdframed}
% [linecolor=gray, backgroundcolor=white, hidealllines, width=0.5]
% \begin{verbatim}
System prompt:\\
You will be provided with answers to questions. Detect the score of $\bigr[$ TRAIT $\bigr]$ for the author of the INPUT from the list [-2, -1, 0, 1, 2] or Nondistinguishable. Use INSTRUCTION.\\
TASK:\\
1. First, list CLUES (i.e., keywords, phrases, contextual information, semantic relations, semantic meaning, tones, references) that support the score determination of $\bigr[$ TRAIT $\bigr]$ of INPUT.\\
2. Second, deduce the diagnostic REASONING process from premises (i.e., clues, input) that supports the INPUT score determination (Limit the number of words to 130).\\
3. Third, based on clues, reasoning and input, determine the score of $\bigr[$ TRAIT $\bigr]$ for the author of INPUT from the list [-2, -1, 0, 1, 2] or Nondistinguishable.\\
4. Mark what made you choose this score as decision type: Explicit signs, Implicit signs, Intuition, Nondistinguishable.\\
5. Provide your output in JSON format with the keys: score, clues, reasoning, decision type.
PROVIDE ONLY JSON.\\
\\
INSTRUCTION:\\
- Definition: $\bigr[$ DEFINITION $\bigr]$\\
- High score of $\bigr[$ TRAIT $\bigr]$ (maximum 2):\\ 
'$\bigr[$ DEFINITION OF HIGH SCORE $\bigr]$'\\
- Low score of $\bigr[$ TRAIT $\bigr]$ (minimum -2):\\ 
'$\bigr[$ DEFINITION OF LOW SCORE $\bigr]$'\\
- Explicit signs: The person mentions obvious facts that are connected with this trait score.\\
- Implicit signs: The person mentions facts that may imply them having this trait score.\\
- Intuition: My intuition tells that the person has this trait score.\\
- Nondistinguishable: I can't tell what trait score the person has.\\
- If the text does not contain substantial, significant, and convincing indicators of the trait score, then use Nondistinguishable.\\
- Choose something other than Nondistinguishable if you have a high degree of confidence in the answer.\\
\\
User prompt:\\
Question: $\bigr[$ QUESTION $\bigr]$\\
INPUT: $\bigr[$ ANSWER $\bigr]$
% \end{verbatim}
\end{mdframed}

\clearpage % Start a new page
\section{Appendix}
\label{lexicon}
\begin{table}[hp!]
\centering
\begin{tabular}{|l|c|c|c|c|c|c|} 
\hline POS & score & Openness &Agreeableness &Conscientiousness & Extraversion & Neuroticism \\ \hline
& high & \begin{tabular}[c]{@{}l@{}}curiosity\\ experience\\ creativity\\ idea\\ world\end{tabular} & \begin{tabular}[c]{@{}l@{}}compassion\\ understanding\\ empathy\\ cooperation\\ kindness\end{tabular} & \begin{tabular}[c]{@{}l@{}}responsibility\\ sense\\ organization\\ attention\\ reliability\end{tabular}   & \begin{tabular}[c]{@{}l@{}}interaction\\ people\\ adventure\\ activity\\ time\end{tabular}      & \begin{tabular}[c]{@{}l@{}}emotion\\ support\\ health\\ stress\\ sense\end{tabular}                              \\ \cline{2-7} 
                            & neutral & \begin{tabular}[c]{@{}l@{}}experience\\ routine\\ hobby\\ place\\ level\\ stability\end{tabular} & \begin{tabular}[c]{@{}l@{}}cooperation\\ work\\ harmony\\ team\\ growth\end{tabular}                  & \begin{tabular}[c]{@{}l@{}}balance\\ structure\\ flexibility\\ organization\\ responsibility\end{tabular} & \begin{tabular}[c]{@{}l@{}}interaction\\ work\\ balance\\ activity\\ solitude\end{tabular}      & \begin{tabular}[c]{@{}l@{}}stability\\ stress\\ challenge\\ cause\\ society\end{tabular}                         \\ \cline{2-7} 
\multirow{3}{*}{NOUN}      & low     & \begin{tabular}[c]{@{}l@{}}stability\\ routine \\ reliability\\ familiarity\\ time\end{tabular}  & \begin{tabular}[c]{@{}l@{}}interest\\ goal\\ relationship\\ success\\ performance\end{tabular}        & \begin{tabular}[c]{@{}l@{}}moment\\ flexibility\\ flow\\ freedom\\ spontaneity\end{tabular}               & \begin{tabular}[c]{@{}l@{}}time\\ friend\\ family\\ introspection\\ interaction\end{tabular}    & \begin{tabular}[c]{@{}l@{}}resilience\\ stability\\ challenge\\ emotion\\ setback\end{tabular}                   \\ \hline
                            & high    & \begin{tabular}[c]{@{}l@{}}explore\\ learn\\ travel\\ embrace\\ pursue\end{tabular}              & \begin{tabular}[c]{@{}l@{}}understand\\ contribute\\ volunteer \\ listen\\ help\end{tabular}          & \begin{tabular}[c]{@{}l@{}}maintain\\ plan\\ prioritize\\ ensure\\ focus\end{tabular}                     & \begin{tabular}[c]{@{}l@{}}engage\\ travel\\ enjoy\\ spend\\ meet\end{tabular}                  & \begin{tabular}[c]{@{}l@{}}provide\\ prioritize\\ promote\\ offer\\ navigate\end{tabular}                        \\ \cline{2-7} 
                            & neutral & \begin{tabular}[c]{@{}l@{}}explore\\ seek\\ travel\\ experience \\ value\end{tabular}            & \begin{tabular}[c]{@{}l@{}}collaborate\\ maintain\\ work \\ strive\\ prioritize\end{tabular}          & \begin{tabular}[c]{@{}l@{}}maintain\\ appreciate\\ detail\\ allow\\ create\end{tabular}                   & \begin{tabular}[c]{@{}l@{}}enjoy\\ maintain\\ appreciate\\ extroverte\\ allow\end{tabular}      & \begin{tabular}[c]{@{}l@{}}volunteer\\ contribute\\ make\\ suggest\\ handle\end{tabular}                         \\ \cline{2-7} 
\multirow{3}{*}{VERB}      & low     & \begin{tabular}[c]{@{}l@{}}establish\\ seek\\ maintain\\ appreciate\\ enjoy\end{tabular}         & \begin{tabular}[c]{@{}l@{}}focus\\ pursue\\ work\\ prioritize\\ achieve\end{tabular}                  & \begin{tabular}[c]{@{}l@{}}live\\ enjoy\\ bogge\\ prefer\\ embrace\end{tabular}                           & \begin{tabular}[c]{@{}l@{}}spend\\ enjoy\\ love\\ work\\ allow\end{tabular}                     & \begin{tabular}[c]{@{}l@{}}bounce\\ compose\\ find\\ remain\\ handle\end{tabular}                                \\ \hline
                            & high    & \begin{tabular}[c]{@{}l@{}}new\\ open\\ different\\ creative\\ innovative\end{tabular}           & \begin{tabular}[c]{@{}l@{}}harmonious\\ positive\\ supportive\\ willing\\ cooperative\end{tabular}    & \begin{tabular}[c]{@{}l@{}}responsible\\ structured\\ reliable\\ strong\\ careful\end{tabular}            & \begin{tabular}[c]{@{}l@{}}social\\ positive\\ dynamic\\ meaningful\\ diverse\end{tabular}      & {\color[HTML]{333333} \begin{tabular}[c]{@{}l@{}}emotional\\ mental\\ complex\\ high\\ challenging\end{tabular}} \\ \cline{2-7} 
                            & neutral & \begin{tabular}[c]{@{}l@{}}new\\ familiar\\ routine\\ balanced\\ open\end{tabular}               & \begin{tabular}[c]{@{}l@{}}personal\\ competitive\\ independent\\ harmonious\\ social\end{tabular}    & \begin{tabular}[c]{@{}l@{}}responsible\\ flexible\\ spontaneous\\ structured\\ manageable\end{tabular}    & \begin{tabular}[c]{@{}l@{}}social\\ independent\\ solitary\\ occasional\\ moderate\end{tabular} & \begin{tabular}[c]{@{}l@{}}emotional\\ passionate\\ stable\\ predictable\\ moderate\end{tabular}                 \\ \cline{2-7} 
\multirow{3}{*}{ADJ.} & low     & \begin{tabular}[c]{@{}l@{}}new\\ familiar\\ adventurous\\ comfortable\\ imaginative\end{tabular} & \begin{tabular}[c]{@{}l@{}}personal\\ focused\\ individual\\ critical\end{tabular}                    & \begin{tabular}[c]{@{}l@{}}spontaneous\\ flexible\\ strict\\ unexpected\\ rigid\end{tabular}              & \begin{tabular}[c]{@{}l@{}}quiet\\ social\\ personal\\ meaningful\\ solitary\end{tabular}       & \begin{tabular}[c]{@{}l@{}}emotional\\ simple\\ present\\ harmonious\\ resilient\end{tabular}                    \\ \hline
\end{tabular}
\end{table}

\end{document}